\documentclass[letterpaper, 10 pt, conference]{ieeeconf}  

\newcommand{\va}{\mathbf{a}}

\newcommand{\vg}{\mathbf{g}}
\newcommand{\vh}{\mathbf{h}}

\newcommand{\vs}{\mathbf{s}}

\newcommand{\vw}{\mathbf{w}}

\newcommand{\mG}{\mathbf{G}}

\newcommand{\mI}{\mathbf{I}}

\newcommand{\vtheta}{\boldsymbol{\theta}}

\newcommand{\vtau}{\boldsymbol{\tau}}

\newcommand{\sE}{\mathbb{E}} 

\newcommand{\sR}{\mathbb{R}} 

\newcommand{\mathA}{\mathcal{A}}

\newcommand{\mathD}{\mathcal{D}} 

\newcommand{\mathH}{\mathcal{H}}

\newcommand{\mathL}{\mathcal{L}} 
\newcommand{\mathM}{\mathcal{M}}
\newcommand{\mathN}{\mathcal{N}}
\newcommand{\mathO}{\mathcal{O}} 

\newcommand{\mathR}{\mathcal{R}}
\newcommand{\mathS}{\mathcal{S}}
\newcommand{\mathT}{\mathcal{T}}

\usepackage{amsmath}
\usepackage{amsfonts}
\usepackage{amssymb}
\usepackage{hyperref}
\usepackage{graphicx}
\usepackage{algorithm}
\usepackage{algorithmic}
\usepackage{mathrsfs}
\usepackage{float}
\usepackage{booktabs} 
\usepackage{color}

\IEEEoverridecommandlockouts                              

\title{\LARGE \bf
Guided Riemannian Optimization (GuRO): Bridging Model Predictive Control and Decision Transformers

}

\author{Hossein Abdi, Satya Prakash Dash, and Mingfei Sun
\thanks{All authors are with the Department of Computer Science, The University of Manchester, Manchester, UK.
}
\thanks{Correspondence:
        {\tt\small {hossein.abdi@manchester.ac.uk}}
}}

\begin{document}

\maketitle
\thispagestyle{empty}
\pagestyle{empty}

\begin{abstract}

Decision-making in high-dimensional, nonlinear systems remains a central challenge in robotics. While model-based methods like Model Predictive Control (MPC) offer sample efficiency and interpretability, their performance degrades when the dynamics model is inaccurate or long-horizon predictions are required. Conversely, model-free reinforcement learning (RL) learns policies directly from interaction but suffers from high sample complexity and unstable optimization. Recent advances in sequence modeling have inspired transformer-based decision-making frameworks that can unify MPC and RL, but their training typically faces significant optimization challenges due to highly non-convex loss landscapes.
In this work, we propose a novel framework that integrates MPC with RL in a sequence decision-making framework and leverages a curvature-aware optimization to efficiently tackle non-convex loss landscapes. MPC provides predictions of locally optimal trajectories that guide the decision transformer, removing the need for extensive offline pretraining. To address the slow and unstable convergence of traditional optimizers, we train the policy in a Riemannian parameter space using an efficient Riemannian (curvature-aware) method, leading to faster and more robust optimization. We evaluate our framework on high-dimensional quadruped control tasks and demonstrate consistent improvements over strong baselines, including TRPO, SAC, and Online Decision Transformer, achieving higher returns and faster convergence.


\end{abstract}

\section{Introduction}

Robotic decision-making with high-dimensional and nonlinear dynamics---such as quadruped locomotion---remains a fundamental and challenging problem in robotics.
While two dominant paradigms, namely model-based control such as \emph{Model Predictive Control (MPC)} \cite{rawlings2020model} and \emph{model-free Reinforcement Learning (RL)} \cite{sutton1998reinforcement}, have each achieved impressive milestones, both have critical limitations.
\textbf{Model-based MPC} is widely adopted for its interpretability and sample efficiency, particularly when an accurate dynamics model is available \cite{mayne2000constrained}. However, the performance degrades when the underlying model is inaccurate and only locally reliable. Moreover, MPC often struggles in tasks that require capturing long-horizon prediction, especially when the approximation of the terminal cost is inaccurate---a scenario commonly observed in complex robotic systems \cite{schwenzer2021review, abdi2024model}.  
\textbf{Model-free RL}, on the contrary, is inherently model-agnostic and can learn complex, high-dimensional policies directly from data through interaction with the environment. However, these methods typically suffer from high sample complexity and unstable optimization dynamics \cite{sutton1998reinforcement}. These limitations motivate the search for policies that can take advantage of both the strengths of model-based reasoning and model-free learning \cite{reiter2025synthesis}.

In recent years, motivated by the successes of natural language processing and computer vision, \emph{sequence modeling} \cite{vaswani2017attention} has been increasingly integrated into both MPC \cite{celestini2024transformer} and RL frameworks \cite{chen2021decision}. This paradigm offers a modern and unified framework for leveraging the complementary strengths of RL and MPC, particularly in high-dimensional robotic systems performing complex tasks that often involve long-term temporal dependencies---scenarios in which conventional methods often face significant limitations. Sequence decision-making approaches, such as Decision Transformer (DT) \cite{chen2021decision} or Trajectory Transformer (TT) \cite{janner2021offline}, formulate policy learning as a sequence modeling problem. Although promising, challenges remain as open research areas in transformer-based policy learning.

Sequence decision-making networks are typically trained on offline datasets, as in offline reinforcement learning. Even approaches designed for online adaptation—such as Online Decision Transformer \cite{zheng2022online}—still rely on extensive pretraining with large offline datasets. This is mainly because purely online training faces non-stationary data distributions, which can destabilize learning and reduce policy performance. Integrating MPC with RL within the sequence modeling framework addresses these challenges by generating locally near-optimal trajectories that guide Decision Transformer learning, thereby mitigating instability from distributional shifts and reducing the need for real-environment interactions, which improves sample efficiency.


Furthermore, training these large transformer-based architectures involves solving a highly non-convex optimization problem, which is often characterized by sharp critical points and complex curvature in the loss manifold. In practice, this problem is typically addressed using standard gradient-based optimization algorithms, such as classical stochastic gradient descent (SGD), Adam \cite{kingma2014adam}, and their variants. However, a key limitation of these methods is that they fail to account for the intrinsic geometry of the loss manifold, including its curvature.
Consequently, these optimizers frequently exhibit slow convergence, high sensitivity to hyperparameter settings, and can produce suboptimal updates or unstable, oscillatory behavior, particularly when hyperparameters are poorly tuned. Moreover, the slower convergence rate increases the sample complexity, as more interactions with the environment are required to achieve comparable performance.

Curvature-aware optimization algorithms, including both Riemannian methods and Euclidean second-order approaches, overcome this limitation by exploiting curvature information to adapt the update direction to the local geometry of the loss landscape. These algorithms are known to mitigate critical slowdowns in optimization \cite{amari1998natural} and can achieve faster convergence compared to traditional gradient-based methods \cite{anil2020scalable}.

To bridge the aforementioned research gaps, we propose a novel approach for policy learning while integrating the model-based MPC with model-free RL. Particularly, our key contributions are as follows:
%
%
\begin{itemize}
    \item Integrating model-based MPC with model-free RL within a sequence decision-making framework. MPC provides real-time, locally optimal, predicted trajectories that enable the training of a decision transformer without requiring offline pretraining.
    \item Employing a Riemannian optimization approach to capture the intrinsic geometry of the loss manifold, which provides fast and stable training of large transformer-based policies.
    \item Validating the proposed framework on high-dimensional quadruped robots across complex control tasks, and conducting extensive comparisons against strong baselines, including TRPO, SAC, and Online-DT. Evaluation metrics include mean total reward and convergence speed.
\end{itemize}

\section{Related Works}

\subsection{Integrating Model Predictive Control with Reinforcement Learning}

The integration of model predictive control and reinforcement learning has been extensively studied from multiple perspectives. A major line of research focuses on learning the underlying Markov decision process structure within model-based RL frameworks. In these approaches, MPC typically operates on an approximate model of the real environment, which is learned iteratively to better capture the system's dynamics or reward function \cite{wu2022plan, chua2018deep}. Some works focus on learning value functions or Q-functions via MPC-based planning \cite{lowrey2018plan, hatch2021value}, while others jointly learn both the dynamics and value functions \cite{hansen2022temporal}.

\noindent
A second stream of research aims to directly learn policies using MPC. In imitation learning, MPC acts as an expert, and neural networks are trained to replicate its behavior for faster online inference \cite{kang2023rl+, mamedov2024safe}. In actor-critic, MPC serves as the actor, and a learned critic evaluates and refines its policy \cite{romero2024actor, gros2019data}. Guided policy search methods represent another main research direction, which leverages MPC-based trajectory optimization to iteratively guide policy learning toward improved solutions \cite{levine2013guided, levine2014learning, carius2020mpc}.

\noindent
Beyond these mainstream approaches, in some works, MPC has been used as a reference generator to provide desired trajectories for RL-based tracking controllers \cite{jenelten2024dtc}, as a safety filter to enforce constraints during exploration \cite{brunke2022safe}, or as a policy generator while RL acts as a policy evaluator \cite{lin2023reinforcement}. In contrast to these existing strategies, our method leverages a model-based MPC to guide a Decision Transformer, which directly optimizes policy-conditioned trajectories.

\subsection{Sequence Modeling for Decision Making}

Transformer-based architectures \cite{vaswani2017attention} have demonstrated remarkable success across natural language processing and computer vision, and their adoption in reinforcement learning has grown rapidly. Decision Transformer (DT) \cite{chen2021decision} and Trajectory Transformer (TT) \cite{janner2021offline} concurrently pioneered reframing policy learning as a sequence modeling problem. They inspired several extensions and variants that address different challenges: Online decision transformer (Online-DT) \cite{zheng2022online} adapts DT for online learning; Constrained DT (CDT) \cite{liu2023constrained} balances safety and task performance; Behavior Transformer (BeT) \cite{shafiullah2022behavior} learns behaviors from multi-modal data; and Bootstrapped Transformer (BooT) \cite{wang2022bootstrapped} improves sequence model training via generating more trajectories. Similarly, Q-learning DT (QDT) \cite{yamagata2023q} leverages dynamic programming to relabel returns, and Pretrained DT (PDT) \cite{xie2023future} introduces Bayesian fine-tuning for online adaptation. Recent works integrate dynamics modeling into DT training, such as \cite{kim2022dynamics}, while the Environment Transformer \cite{wang2024environment} facilitates model-based RL by learning environment dynamics jointly with policy representations.

\noindent
In parallel, transformer-based sequence modeling has also been explored in the context of MPC. Transformers have been used to warm-start optimal control problems \cite{celestini2024transformer}; to learn dynamic models for MPC \cite{kotb2024qt}; and, more recently, M$^3$PC \cite{citation-0} integrates MPC into a bidirectional trajectory model to improve action selection at test time.
Our approach builds on these ideas but differs fundamentally: we employ model-based MPC to guide a Decision Transformer during training policy-conditioned trajectories through a Riemannian optimization framework.

\subsection{Curvature-aware optimization algorithms}

Curvature-aware optimization has long been studied for improving convergence in high-dimensional learning problems.
From a Euclidean perspective, classical second-order methods such as Newton’s method and quasi-Newton schemes (e.g., BFGS) \cite{nocedal2006numerical} exploit the Hessian matrix to precondition gradients, thus adapting the update step to the local curvature of the loss landscape.
From a Riemannian perspective, Natural Gradient Descent (NGD) \cite{amari1998natural, dash2026gradient} interprets the parameter space as a Riemannian manifold and leverages the Fisher information matrix as a local Riemannian metric to capture the loss curvature, a method used in the natural policy gradient \cite{kakade2001natural} within the policy optimization.

\noindent
However, despite their theoretical advantages, the practical deployment of these methods in high-dimensional policy learning is often constrained by their computational and memory costs. 
To overcome these challenges, a wide range of approximation techniques has been developed. For instance, K-FAC \cite{martens2015optimizing} exploits Kronecker-factored approximations of block-diagonal Fisher matrices. Shampoo \cite{gupta2018shampoo} maintains a set of per-dimension preconditioning matrices that approximate curvature in a scalable manner, while AdaHessian \cite{yao2021adahessian} and Sophia \cite{liu2023sophia} adopt a diagonal Hessian approximation to reduce computational overhead. Alternatively, conjugate gradient methods \cite{nocedal2006numerical} bypass explicit preconditioning by directly solving the approximate natural gradient step in Riemannian space, a strategy notably employed by TRPO \cite{schulman2015trust}.

\section{Preliminaries and Background}

\subsection{Problem Statement}
\label{subsec:problem_statement}

We consider a sequential decision-making problem that is mathematically formulated within the Markov Decision Process (MDP) framework \cite{bellman1957markovian}. Formally, an MDP is defined as $\mathM = \langle \mathS, \mathA, P, R, \rho, \gamma \rangle$, where $\mathS$ denotes the state space and $\mathA$ the action space. The transition dynamics is specified by $P(\vs_{t+1} \mid \vs_t, \va_t)$, where $\vs \in \mathS  \subseteq \sR^{d_s}$ and $\va \in \mathA  \subseteq \sR^{d_a}$ represent the state and action vectors, respectively. The reward function is defined as $R(\vs,\va): \mathS \times \mathA \rightarrow \sR$, with $r_t = R(\vs_t, \va_t)$ denoting the instantaneous reward. The distribution $\rho$ specifies the initial state distribution and $\gamma \in [0,1]$ is the discount factor controlling the trade-off between immediate and future rewards. 

\noindent
An agent starts from an initial state $\vs_0 \sim \rho$ and interacts with the environment by sampling actions from a policy $\pi$,
i.e., $\va_t \sim \pi(\cdot \mid \vs_t)$. More generally, the policy may also be conditioned on the entire history of observations,
$\va_t \sim \pi(\cdot \mid \vh_t)$. 
After executing an action $\va_t$, the agent receives a reward $r_t$ and transitions
to the next state $\vs_{t+1} \sim P(\cdot \mid \vs_t, \va_t)$.
The objective is to find the optimal policy $\pi^*$ from a family of policies parameterized by $\vtheta \in \sR^n$ (denoted $\pi_{\theta}$), that maximizes the expected discounted return: 

\begin{equation}
    \pi^* = \underset{\pi_{\theta}}{\mathrm{arg \; max}} \; \sE_{\pi_{\theta}} \bigg[ \sum_{t=0}^{\infty} \gamma^t r_t \bigg]
\end{equation}



\subsection{Model Predictive Control (MPC)}

Consider the following stochastic, nonlinear, discrete-time dynamic model:
\begin{equation}
    \label{eq:stochastic_dynamic_model}
    \vs_{t+1} = f(\vs_t,\va_t) + \mathsf{\vw}_t,
\end{equation}
where $f(\vs_t,\va_t)$ indicates the nominal model, and $\mathsf{\vw}_t \sim \mathN(0,\Sigma_{\mathsf{w}})$ denotes white noise, assumed to be independently and identically distributed (i.i.d.) Gaussian with zero mean and covariance $\Sigma_{\mathsf{w}}$.

%
\noindent
Formally, MPC determines the optimal sequence of control actions by minimizing a cost function over a finite prediction horizon, formulated as the following optimization problem:


\begin{subequations} \label{eq:mpc}
    \begin{align} 
         \underset{\va_{t:t+T|t}}{\mathrm{min}} \; \sE_{\mathsf{\vw}_{t:t+T}} \bigg[ \sum_{i=0}^{T} \ell\!\left(\vs_{t+i|t},\va_{t+i|t}\right) & + \ell_{T+1}\!\left(\vs_{t+T+1|t}\right) \bigg]
         \label{eq:mpc_cost}\\
        \vs_{t+i+1|t} \sim 
        P(\cdot|\vs_{t+i|t}, \va_{t+i|t}), & \quad i= 0,\hdots,T
        \label{eq:mpc_model}\\
        \vs_{t|t} = \vs_t, &
        \label{eq:mpc_state_constrain} 
    \end{align}
\end{subequations}
where $\ell\!\left(\vs_{t+i|t},\va_{t+i|t}\right)$ and $\ell_{T+1}\!\left(\vs_{t+T+1|t}\right)$ denote the stage and terminal cost functions, respectively. Here, $P(\cdot|\vs_{t+i|t}, \va_{t+i|t}) = \mathN\!\left(f(\vs_{t+i|t},\va_{t+i|t}),\Sigma_{\mathsf{w}}\right)$ describes the stochastic dynamic model as defined in \eqref{eq:stochastic_dynamic_model}. Also, $\vs_{t+i|t}$ and $\va_{t+i|t}$ represent the $i^{th}$ predicted state and control action conditioned on the observed state at time-step $t$. This optimization problem is solved using a standard trajectory optimization algorithm, which produces the optimal sequence of control actions: $\va^*_{t:t+T|t}$. For notational simplicity, we omit the superscript $\ast$ when referring to the output of MPC in the remainder of the paper.

\subsection{Riemannian Optimization}

Formally, let $\big(\mathscr{M}, \mG(\vtheta)\big)$ denote an $n$-dimensional Riemannian manifold that represents the parameter space, where $\mathscr{M}$ is a smooth differentiable manifold and $\mG(\vtheta)$ is a Riemannian metric tensor defined at each point $\vtheta \in \mathscr{M}$.
Given a smooth objective function $\mathL: \mathscr{M} \to \sR$, the goal of Riemannian optimization is to solve the following problem:
\begin{equation}
\label{eq:riemannian_optimization}
    \vtheta^\ast = \underset{\vtheta \in \mathscr{M}} {\mathrm{arg \; min}} \;
    \mathL(\vtheta).
\end{equation}
To this end, the Riemannian steepest descent method updates the parameters at the $k^{th}$ iteration ($\vtheta_k$) according to:
\begin{equation}
\label{eq:riemannian_update}
    \vtheta_{k+1} = \mathrm{Retr}_{\theta_k}\big(-\eta_{\theta} \cdot \mathrm{grad} \mathL(\vtheta_k)\big),
\end{equation}
where $\mathrm{grad} \mathL(\vtheta_k)$ denotes the Riemannian gradient of $\mathL(\vtheta)$ at $\vtheta_k$ with respect to the Riemannian metric $\mG(\vtheta)$, $\eta_{\theta}>0$ is the step-size (or learning rate), and $\mathrm{Retr}_{\theta}: \mathscr{T}_{\theta}\mathscr{M} \to \mathscr{M}$ is a retraction mapping that projects a point from the tangent space $\mathscr{T}_{\theta}\mathscr{M}$ back onto the manifold $\mathscr{M}$.
For a rigorous treatment of the underlying mathematical concepts and a comprehensive review of Riemannian optimization techniques, we refer the reader to \cite{sato2021riemannian}.

\section{Method}

\subsection{Sequence Modeling Setup}
Given an MDP $\mathM = \langle \mathS, \mathA, P, R, \rho, \gamma \rangle$ as defined in Section \ref{subsec:problem_statement}, the agent starts the sequential interaction with the environment from an initial state and produces a sequence of states, actions, and rewards—i.e., a trajectory. Following \cite{chen2021decision}, we replace rewards with the return-to-go, which yields the trajectory representation used for autoregressive training and generation: $\vtau_{0:t} = (\vs_0,\va_0,\mathR_0, \vs_1,\va_1,\mathR_1, ..., \vs_t,\va_t,\mathR_t)$. Here, return-to-go is defined as $\mathR_i = \sum_{j=i}^t \gamma^{j-i} r_{j}$, where $i = 0,1, \dots, t$.
The trajectory $\vtau_{0:t}$ is then provided as input to the DT, which outputs the policy distribution $\pi_{\theta}(\cdot|\vtau_{0:t})$.





\subsection{Guiding by MPC}
Traditional decision transformers trained exclusively on offline datasets often perform sub-optimally due to the inherent non-stationarity of data distributions in complex environments. To address this, we incorporate online trajectory generation using model predictive control. Specifically, at each time step during rollout, we execute a finite-horizon MPC with horizon $T$, which serves as the behavior policy to predict an optimal sequence of control actions, $\va_{0:T}$, along with the corresponding state and stage-cost sequences (where the stage cost is equivalent to the negative reward). Together, these outputs define a locally optimal trajectory, $\vtau_{0:T}$, which is then stored in a replay buffer alongside real rollout trajectories, $\mathD = \{\vtau^{(1)}_{0:T}, \vtau^{(2)}_{0:T}, ...\}$, and used for subsequent training of the decision transformer $\pi_{\theta}(\va|\vtau)$.

%
\noindent
To estimate the parameters of the stochastic policy $\pi_{\theta}(\va|\vtau)$ via maximum likelihood estimation (MLE), we minimize the negative log-likelihood (NLL) loss over a mini-batch of trajectories sampled from the replay buffer $\mathD$:
%
%
\begin{equation}
    \mathL_\text{NLL} = \sE_{\vtau \sim \mathT} \bigg[ - \frac{1}{T} \sum_{t=1}^T \log \pi_{\theta}(\va_t | \vtau_{0:t}) \bigg],
\end{equation}
where $\mathT$ denotes the trajectory distribution induced by the replay buffer, and we approximate the expectation $\sE_{\vtau \sim \mathT}$ by averaging over the mini-batch of sampled trajectories.
Optimizing this objective alone often leads to suboptimal, near-deterministic policies and can induce instability due to unbounded updates in the policy parameters. To mitigate these issues, we augment optimization with two complementary constraints:
maximum entropy principle and a Kullback–Leibler (KL) divergence trust region constraint.
The maximum entropy principle encourages policies to maintain stochasticity, thereby promoting sufficient exploration of the action space and avoiding premature convergence to suboptimal deterministic policies \cite{haarnoja2018soft}. 
Simultaneously, the KL-divergence constraint restricts the magnitude of policy updates between successive iterations, which ensures monotonic policy improvement and stabilizes training \cite{schulman2015trust}. Specifically, we optimize:
%
\begin{subequations} \label{eq:nll-kl-entropy}
    \begin{align} 
        \underset{\vtheta}{\mathrm{min}} \;  \sE_{\vtau \sim \mathT}  \bigg[ - \frac{1}{T} & \sum_{t=1}^T \log \pi_{\theta}(\va_t | \vtau_{0:t}) \bigg]
         \label{eq:nll_cost}\\
        \mathrm{s.t.} \nonumber \\
        \sE_{\vtau \sim \mathT} \big[\mathH \big(\pi_{\theta}& (\cdot | \vtau) \big) \big] \geq \delta_{\mathH},
        \label{eq:entopy_const}\\
        \sE_{\vtau \sim \mathT}  \big[D_{\mathrm{KL}}\big(\pi_\mathrm{ref}&(\cdot|\vtau)|\pi_{\theta}(\cdot|\vtau)\big)\big] \leq \delta_\mathrm{KL}, 
        \label{eq:kl_const} 
    \end{align}
\end{subequations}
where $\mathH \big(\pi_{\theta} (\cdot | \vtau) \big) = - \int_{\mathA} \pi_{\theta}(\va|\vtau) \log \pi_{\theta}(\va|\vtau) d\va$ represents differential entropy of policy $\pi_{\theta}$ conditioned on trajectory $\vtau$.
And, $D_{\mathrm{KL}}\big(\pi_\mathrm{ref}(\cdot|\vtau)|\pi_{\theta}(\cdot|\vtau)\big) = \int_{\mathA} \pi_\mathrm{ref}(\va|\vtau) \log \frac{\pi_\mathrm{ref}(\va|\vtau)}{\pi_{\theta}(\va|\vtau)} d\va$ indicates the KL divergence of the reference policy distribution $\pi_\mathrm{ref}$ from the current policy distribution $\pi_{\theta}$.
Here, $\delta_{\mathH}$ denotes the entropy lower bound required to encourage sufficient exploration, and $\delta_\mathrm{KL}$ specifies the trust region bound that limits policy deviation from the reference policy $\pi_\mathrm{ref}$ (i.e., the policy from the previous iteration).
%

\noindent
To solve the constrained optimization problem in \eqref{eq:nll-kl-entropy} in practice, we reformulate it as a primal-dual optimization problem followed by a backtracking line search.
Specifically, we incorporate the hard entropy constraint as a regularization term in the objective and enforce the KL divergence constraint using the backtracking line search technique to stay within the trust region. This yields a primal-dual optimization problem with the primal formulation as follows:
\begin{subequations} \label{eq:nll_regularized_kl}
    \begin{align} 
         \underset{\vtheta}{\mathrm{min}} \;  \sE_{\vtau \sim \mathT}  \Bigg[& \frac{1}{T} \sum_{t=1}^T \bigg( -\log \pi_{\theta}(\va_t | \vtau_{0:t}) \notag \\
         &+ \alpha \Big(\delta_{\mathH} - \mathH \big(\pi_{\theta} (\cdot | \vtau_{0:t}) \big)\Big) \bigg)  \Bigg]
         \label{eq:nll_regularized} \\
        &\mathrm{s.t.} \nonumber \\
        \sE_{\vtau \sim \mathT}  \big[D_{\mathrm{KL}}\big(&\pi_\mathrm{ref}(\cdot|\vtau)|\pi_{\theta}(\cdot|\vtau)\big)\big] \leq \delta_\mathrm{KL}, 
        \label{eq:regularized_kl_const} 
    \end{align}
\end{subequations}
where $\alpha > 0$ serves as the dual variable associated with the entropy constraint.

\noindent
To solve the dual problem, the gradient descent method is used in a log-space ($\beta = \log \alpha$) to ensure that the dual variable $\alpha$ remains positive:
\begin{align} \label{eq:dual}
{\small
    \beta_{k} \leftarrow \beta_{k-1} - \eta_{\beta} \nabla_\beta e^{\beta_{k-1}} \sE_{\vtau \sim \mathT}\big[ \frac{1}{T} \sum_{t=1}^T \mathH \big(\pi_{\theta} (\cdot | \vtau_{0:t}) \big) - \delta_{\mathH}\big].
    }
\end{align}
Here, $\eta_{\beta}>0$ is the learning rate, and $k$ represents the iteration. Moreover, in practice, the expectation $\sE_{\vtau \sim \mathT}$ is approximated empirically using a mini-batch of sampled trajectories.

\subsection{Efficient Riemannian Optimization}

This section details how we leverage an efficient Riemannian optimization algorithm to exploit the underlying curvature of the loss landscape associated with the non-convex primal optimization problem defined in \eqref{eq:nll_regularized_kl}. By accounting for the local geometry, the proposed method enables more informed update steps, thereby improving the per-iteration progress toward the optimum. This, in turn, accelerates convergence and reduces the required number of environment interactions, ultimately reducing the overall complexity of the sample.

\noindent
Formally, we seek an update direction $\delta\vtheta$ that minimizes the loss function, $\mathL(\vtheta + \delta\vtheta)$, while ensuring that the resulting change in the loss, $\Delta \mathL$, corresponds to the steepest descent direction in the induced Riemannian geometry. This is achieved by incorporating the local curvature information, as encoded by an appropriate Riemannian metric, into the optimization process \cite{amari2000methods}.

\noindent
Recall the Riemannian optimization problem defined in \eqref{eq:riemannian_optimization}, which seeks a minimizer $\vtheta^\ast$ on a smooth manifold $\mathscr{M}$. In our policy learning setting, the optimization problem in \eqref{eq:nll_regularized} can be interpreted as a special case of Riemannian formulation, where the Euclidean space $\sR^n$ with the standard inner product can be naturally regarded as a Riemannian manifold $\mathscr{M} \equiv \sR^n$ \cite{sato2021riemannian}.

\noindent
Building on this reformulation with the natural retraction $\mathrm{Retr}_{\theta}(\nu)=\vtheta+\nu$, the update direction for the policy parameters at iteration $k$ corresponds to the steepest descent direction in the Riemannian space. Specifically, the parameter update is given by
\begin{equation}
\label{eq:riemannian_update_direction}
    \delta\vtheta_k = -\eta_{\theta} \cdot \mathrm{grad} \mathL(\vtheta_k),
\end{equation}

\noindent
where the Riemannian gradient $\mathrm{grad} \mathL(\vtheta_k)$ with respect to the Riemannian metric $\mG(\vtheta_k)$ is defined as:
\begin{equation}
\label{eq:riemannian_grad}
\mathrm{grad} \mathL(\vtheta_k) = \mG(\vtheta_k)^{\dagger} \nabla_{\theta} \mathcal{L}(\vtheta_k).
\end{equation}
Here, $\mG(\vtheta_k)^{\dagger}$ denotes the Moore–Penrose pseudo-inverse of the Riemannian metric tensor $\mG(\vtheta_k)$. The metric tensor encodes the local geometric structure of the parameter space and, consequently, the curvature of the loss landscape. In this formulation, $\mG(\vtheta_k)^{\dagger}$ acts as a preconditioner for the Euclidean gradient to rescale and reorient the update direction toward the steepest descent direction in the induced Riemannian geometry.


\begin{algorithm}[!t]
\caption{GuRO: Guided Riemannian Optimization for Decision Transformer learning via MPC.}
\label{alg:mpc_rl_dt}
\begin{algorithmic}[1]
    \STATE \textbf{Input:} primal step-size $\eta_{\theta}$; dual learning rate $\eta_{\beta}$; forgetting factor $\lambda$; batch size $m$; MPC horizon $T$; discount factor $\gamma$; entropy lower bound $\delta_{\mathH}$; trust region bound $\delta_\mathrm{KL}$; contraction factor $c$; epsilon value $\epsilon$.
    \STATE \textbf{Initialize:} policy parameters $\vtheta_0$; dual variable $\alpha_0$; Riemannian metric estimate $\bar{\mG}_0 = \mathO$; replay buffer $\mathD$.
    \STATE \textbf{Output:} optimal policy $\pi^\ast(\va|\vtau) = \pi_{\theta^\ast}(\va|\vtau)$
    \FOR{$k = 1, 2, \dots$}
        \STATE $\{\vtau^{(1)}_{0:T}, \vtau^{(2)}_{0:T}, \dots\} \gets$ Rollout and Run MPC.
        \STATE $\mathD \gets \mathD \cup \{\vtau^{(1)}_{0:T}, \vtau^{(2)}_{0:T}, \dots\}$
        \STATE $ \{\vtau^{(b)}_{0:T}\}^{m}_{b=1} \gets$ Sample batch of trajectories from $\mathD$.
        \STATE $\mathH \big(\pi_{\theta} (\cdot | \vtau^{(b)}_{0:t}) \big)$ \hfill  $\triangleright$  $\forall\, b=1,\dots,m$ and $\forall\, t=1,\dots,T$
        \STATE Update $\alpha_k$ \hfill $\triangleright$ Equation \eqref{eq:dual}, where $\beta = \log \alpha$.
        \STATE $ \hat{\va}^{(b)}_{t} {\scriptstyle\sim} \pi_{\theta}(\cdot|\vtau^{(b)}_{0:t})$   $\triangleright$  $\forall\, b=1,\dots,m$ and $\forall\, t=1,\dots,T$
        \STATE Compute gradient estimate: $\hat{\vg}$ \hfill $\triangleright$ Equation \eqref{eq:gradient} 
        \STATE Estimate $\hat{\mG}$ \hfill $\triangleright$ Equation \eqref{eq:instantaneous_riemannian}
        \STATE Update $\bar{\mG}_k$ \hfill $\triangleright$ Equation \eqref{eq:ema_riemannian}
        \STATE Euclidean gradient: $\nabla_{\theta} \mathL(\vtheta_k)$ \hfill $\triangleright$ Equation \eqref{eq:euclidean_gradient} 
        \STATE Riemannian gradient: $\mathrm{grad} \mathL(\vtheta_k) = \bar{\mG}_{k}^{\dagger} \odot \nabla_{\theta} \mathcal{L}(\vtheta_k)$ 
        \STATE \textbf{Trust Region Backtracking Line Search:}
        \STATE Re-initialize step-size $\eta_{\theta}$
        \REPEAT
            \STATE Update $\vtheta_k$ along $\mathrm{grad} \mathL(\vtheta_k)$ \hfill $\triangleright$ Equation \eqref{eq:covariant}
            \STATE Estimate $\hat{D}_{\mathrm{KL}}\big(\pi_{\theta_{k-1}}|\pi_{\theta_{k}}\big)$ \hfill $\triangleright$ Equation \eqref{eq:mini_batch_kl}
            \STATE $\eta_{\theta} \leftarrow c \cdot \eta_{\theta}$ \hfill $\triangleright$ Shrink step size.
        \UNTIL{$\hat{D}_{\mathrm{KL}}\big(\pi_{\theta_{k-1}}|\pi_{\theta_{k}}\big) \leq \delta_\mathrm{KL}$}
    \ENDFOR
    \STATE \textbf{return} Optimal parameters $\vtheta^\ast$ 
\end{algorithmic}
\end{algorithm}

\noindent
A key challenge, however, lies in computing the Riemannian metric $\mG(\vtheta)$. Direct computation of $\mG(\vtheta)$ is generally intractable for high-dimensional policies, such as decision transformers, due to the prohibitive computational and memory costs associated with exact Riemannian metric formulations. To overcome this limitation, we propose a scalable, mini-batch-based approximation strategy that leverages empirical expectations, exponential moving averages, and diagonalized tensor representations.

\noindent
Following \cite{pascanu2013revisiting}, we approximate the diagonal Riemannian metric using a Gauss–Newton estimator. Concretely, we sample a mini-batch of size $m$ from the replay buffer $\mathD$,
yielding trajectories $ \{\vtau^{(b)}_{0:T}\}^{m}_{b=1}$.
For each trajectory sample $b \in \{1, \dots, m\}$ and time steps $t \in \{1, \dots, T\}$, we draw actions from the current policy distribution: $ \hat{\va}^{(b)}_{t} \sim \pi_{\theta}(\cdot|\vtau^{(b)}_{0:t})$. Using these samples, we compute the mini-batch gradient estimate of the negative log-likelihood:
\begin{equation}
\label{eq:gradient}
    \hat{\vg} = -\frac{1}{mT} \sum_{b=1}^m \sum_{t=1}^T \nabla_{\theta} \log \pi_{\theta}(\hat{\va}^{(b)}_{t} | \vtau^{(b)}_{0:t}) .
\end{equation}
%
The corresponding instantaneous diagonal Riemannian metric is then defined as \cite{pascanu2013revisiting}:
\begin{equation}
\label{eq:instantaneous_riemannian}
    \hat{\mG} = m\;T \; \mathrm{diag}(\hat{\vg} \cdot \hat{\vg}^\top+\epsilon \mI_n),
\end{equation}
where the function $\mathrm{diag}(\cdot)$ extracts the diagonal entries of its matrix argument, and $0 < \epsilon \ll 1$ is for stability. Here, $\mI_n$ denotes the $n \times n$ identity matrix matching the parameter dimensionality.

\noindent
We then maintain an exponential moving average to obtain a stable and low-variance approximation of the metric:
\begin{equation}
\label{eq:ema_riemannian}
    \bar{\mG}_k = \lambda \bar{\mG}_{k-1} + (1-\lambda) \hat{\mG},
\end{equation}
where $\bar{\mG}_k$ denotes the smoothed Riemannian metric approximated at iteration $k$, and $\lambda \in (0,1)$ is the forgetting factor that controls the trade-off between stability and adaptability.
Since $\bar{\mG}_k$ stores only the diagonal entries of the original matrix, its Moore–Penrose pseudo-inverse $\bar{\mG}_{k}^{\dagger}$ is obtained by inverting these entries elementwise.

\noindent
Next, using the sampled trajectories, we calculate the mini-batch Euclidean gradient of the entropy-regularized loss in \eqref{eq:nll_regularized} as below:
\begin{align}
\label{eq:euclidean_gradient}
\nabla_{\theta} \mathL(\vtheta) = \frac{1}{mT} \sum_{b=1}^m \sum_{t=1}^T &\nabla_{\theta} \bigg[ -\log \pi_{\theta}(\va^{(b)}_t | \vtau^{(b)}_{0:t}) \notag \\
         + &\alpha \Big(\delta_{\mathH} - \mathH \big(\pi_{\theta} (\cdot | \vtau^{(b)}_{0:t}) \big)\Big) \bigg].
\end{align}
%
%
%
%
Given $\bar{\mG}_{k}^{\dagger}$ and $\nabla_{\theta} \mathL(\vtheta)$, we approximate the Riemannian gradient according to \eqref{eq:riemannian_grad}. Consequently, the primal optimization problem defined in \eqref{eq:nll_regularized_kl} can be solved iteratively in the Riemannian parameter space based on the approximated diagonal metric. Specifically, at iteration $k$, the policy parameters are updated via:
\begin{align}
\label{eq:covariant}
    \vtheta_k \leftarrow \vtheta_{k-1} - \eta_{\theta} \bar{\mG}_{k-1}^{\dagger} \odot \nabla_{\theta} \mathcal{L}(\vtheta_{k-1}),
\end{align}
where $\odot$ denotes the element-wise (Hadamard) product.
We then adapt the step-size $\eta_{\theta}$ using a contraction factor $c \in (0,1)$ by performing a backtracking line search along the direction $\delta\vtheta_k = -\eta_{\theta} \bar{\mG}_{k}^{\dagger} \odot \nabla_{\theta} \mathcal{L}(\vtheta_k)$ to ensure monotonic improvement of the objective while satisfying the trust region constraint in \eqref{eq:regularized_kl_const}. To this aim, the mini-batch estimate of the KL divergence is computed as:

\begin{align}
\label{eq:mini_batch_kl}
{\small
\hat{D}_{\mathrm{KL}}\big(\pi_\mathrm{ref}|\pi_{\theta}\big) = \frac{1}{mT} \sum_{b=1}^m \sum_{t=1}^T D_{\mathrm{KL}}\big(\pi_{\theta_{k-1}}(\cdot|\vtau^{(b)}_{0:t})|\pi_{\theta_{k}}(\cdot|\vtau^{(b)}_{0:t})\big).
}
\end{align}

\noindent
The complete procedure is outlined in Algorithm~\ref{alg:mpc_rl_dt}.

\section{Experiments and Discussion}

We evaluate the effectiveness of the proposed algorithm on a set of high-dimensional, nonlinear robotic control tasks, focusing on quadruped locomotion in challenging environments.
\subsection{Validation Setup}
\begin{figure}[!t]
\vspace{0.2cm}
  \centering
 \includegraphics[width=1.0\linewidth]{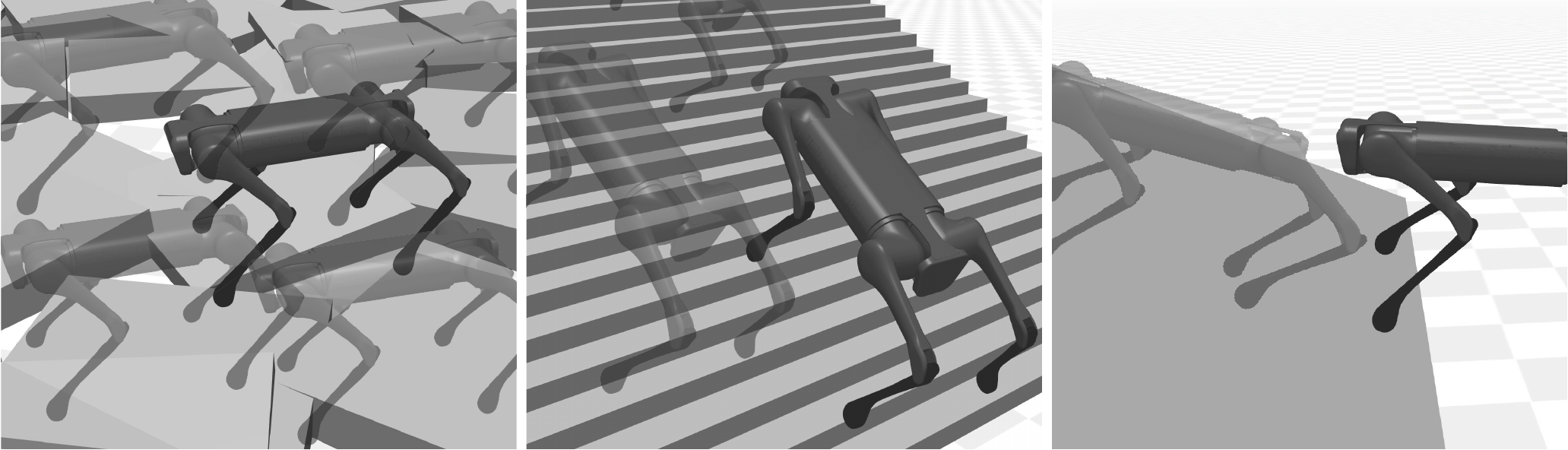}
  \caption{Experimental evaluation on the \textbf{Unitree AlienGo} quadruped robot, simulated using the MuJoCo XLA (MJX) physics engine within the JAX framework. All experiments leverage a parallelized GPU implementation on an NVIDIA GeForce RTX 4090. The translucent robots illustrate the parallel simulations running simultaneously. The tasks include \emph{locomotion on uneven terrains} (left), \emph{stair climbing} (center), and \emph{ascending a low-friction inclined surface} (right).}
  \label{fig:robot}
\end{figure}

\textbf{Environments and Tasks.} 
All experiments are conducted using the \emph{Unitree AlienGo} quadruped robot, evaluated on three distinct challenging locomotion scenarios: (i) locomotion on uneven terrains, 
(ii) stair climbing, and
(iii) ascending a low-friction inclined surface--as illustrated in Figure \ref{fig:robot}.
%

\textbf{Baselines and Metrics.}
We compare our method against a diverse set of competitive baselines, including state-of-the-art reinforcement learning algorithm: Soft Actor-Critic (SAC) \cite{haarnoja2018soft}; curvature-aware policy optimization method: Trust Region Policy Optimization (TRPO) \cite{schulman2015trust}; and online variant of the decision transformer: Online-DT \cite{zheng2022online}.
To rigorously assess the contribution of our Riemannian optimization algorithm on MPC-guided DT learning, we additionally evaluate its performance compared with widely used Euclidean optimizers: SGD and Adam.
Performance is measured in terms of average total reward, sample efficiency, and convergence rate, which provides a comprehensive assessment of both learning speed and final policy quality.

\textbf{Implementation Settings.}
The experiments are performed on the Unitree AlienGo quadruped robot, simulated using the MuJoCo XLA (MJX) physics engine within the JAX framework. All experiments leverage a parallelized GPU implementation on an NVIDIA GeForce RTX 4090, and the average and standard deviation over 5 runs with different initial random seeds are reported.
As a Decision Transformer, we employ a GPT-style sequence model implemented as a 3-layer Transformer decoder with 128-dimensional hidden states and 512-dimensional feedforward layers.
For the MPC component, we employ the whole-body dynamics model as the nominal dynamics, assuming an uncertainty with a standard deviation of $0.01$ in each dimension.
The quantities $\mathH \big(\pi_{\theta} (\cdot | \vtau) \big)$ and $D_{\mathrm{KL}}\big(\pi_\mathrm{ref}(\cdot|\vtau)|\pi_{\theta}(\cdot|\vtau)\big)$ in the Equations \eqref{eq:dual}, \eqref{eq:euclidean_gradient}, and \eqref{eq:mini_batch_kl} are computed using the built-in implementations available in PyTorch.

\begin{figure}[H]
\vspace{0.2cm}
  \centering
  \includegraphics[width=1.0\linewidth]{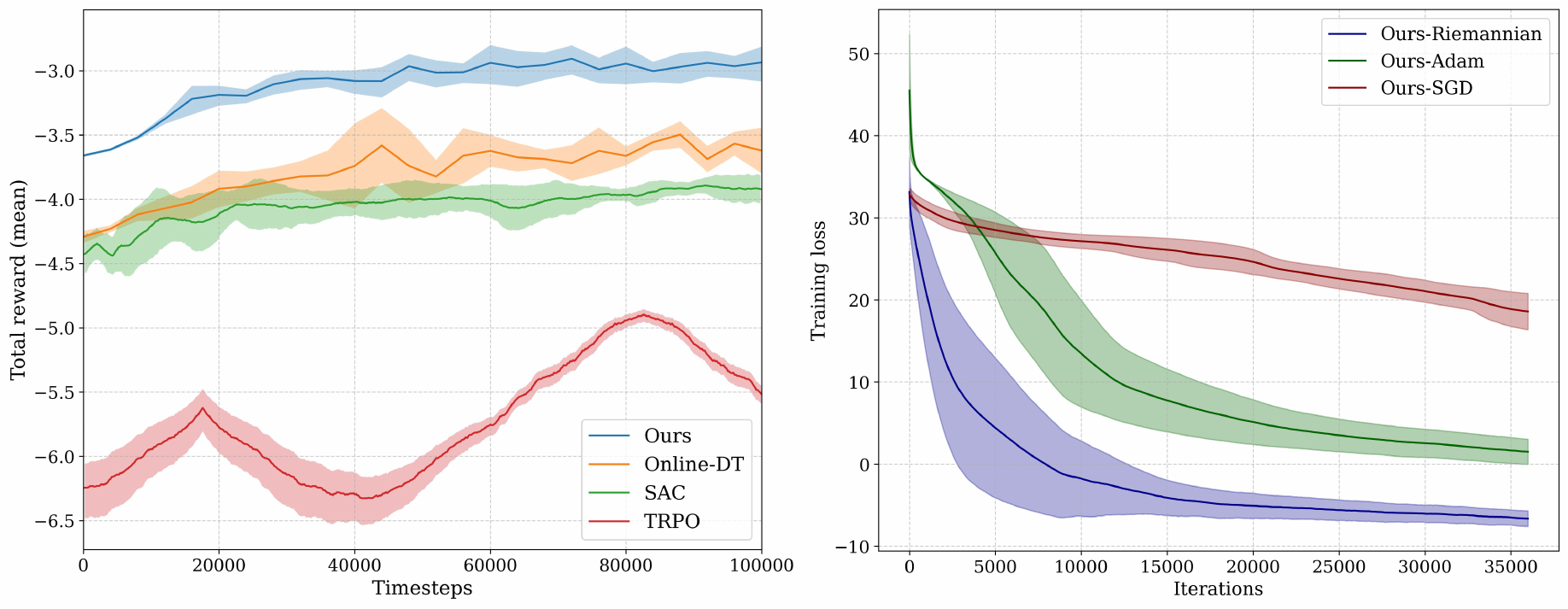}
  \vspace{-0.5cm}
  \caption{\textbf{Locomotion on uneven terrains.} \textit{Left}: mean total reward versus timesteps, highlighting improved sample efficiency and final performance. \textit{Right}: training loss across iterations, illustrating faster convergence with the proposed Riemannian optimization method.}
  \label{fig:robot_neven_terrain}
\end{figure}
\vspace{-0.4cm}

\begin{figure}[H]
  \centering
  \includegraphics[width=1.0\linewidth]{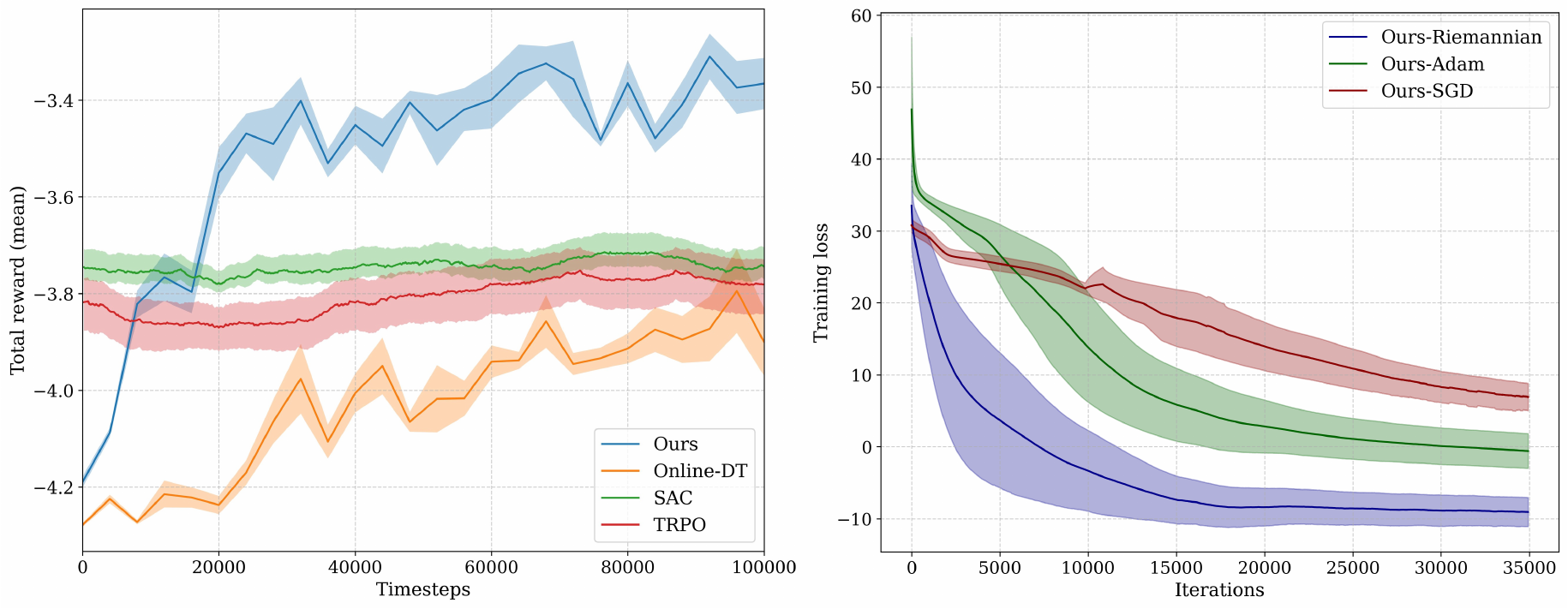}
  \vspace{-0.5cm}
  \caption{\textbf{Stair climbing performance.} \textit{Left}: mean total reward over timesteps, showing enhanced sample efficiency and final reward with our approach. \textit{Right}: corresponding training loss, demonstrating accelerated convergence relative to SGD and Adam.}
  \label{fig:robot_stair_climbing}
\end{figure}
\vspace{-0.4cm}

\begin{figure}[H]
  \centering
  \includegraphics[width=1.0\linewidth]{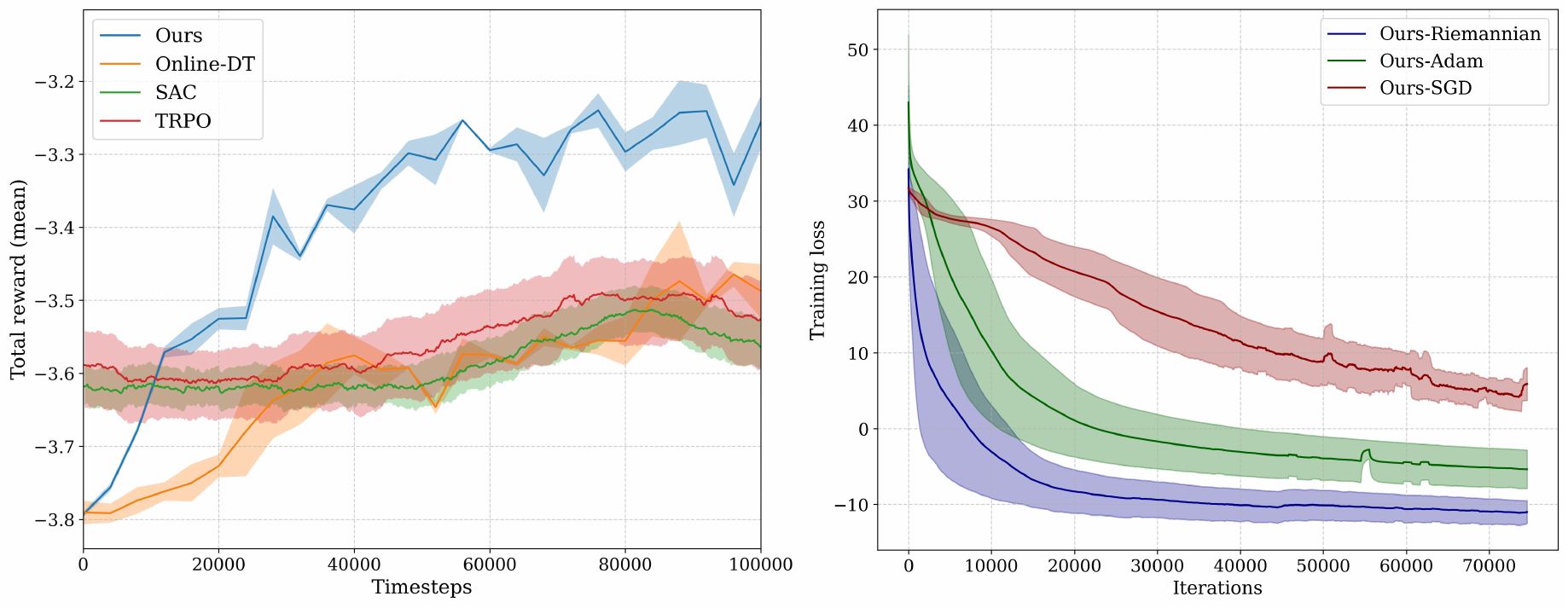}
  \vspace{-0.5cm}
  \caption{\textbf{Ascending low-friction incline.} \textit{Left}: mean total reward, emphasizing the performance improvement and sample efficiency of our method. \textit{Right}: training loss across iterations, highlighting the faster convergence of the proposed Riemannian optimizer compared to conventional methods.}
  \label{fig:robot_inclined_surface}
\end{figure}

\begin{table*}
\vspace{0.2cm}
  \small
  \caption{Average total reward (mean $\pm$ standard deviation) for quadruped robot across challenging scenarios. Tasks include uneven terrain locomotion, stair climbing, and low-friction slope ascent. Higher values correspond to better performance.}
  \label{table:result}
  \centering
  {\small
  \begin{tabular}{c c c c c c c}
    \toprule
    \textbf{Task Scenario} & \textbf{TRPO} &
    \textbf{SAC} & \textbf{Online-DT} & \textbf{Ours-SGD} & \textbf{Ours-Adam} & \textbf{Ours-Riemannian}  \\
    \midrule
    \textbf{Locomotion on uneven terrains} & $-5.52_{\pm0.2}$  & $-3.85_{\pm0.2}$ & $-3.65_{\pm0.3}$ & $-3.42_{\pm0.4}$ & $-3.25_{\pm0.3}$ & $\boldsymbol{-2.95_{\pm0.2}}$  \\ \addlinespace[1ex]
    \textbf{Stair climbing performance}      & $-3.78_{\pm0.4}$ & $-3.68_{\pm0.3}$ & $-3.90_{\pm0.4}$ & $-3.52_{\pm0.4}$ & $-3.42_{\pm0.3}$ & $\boldsymbol{-3.36_{\pm0.2}}$  \\ \addlinespace[1ex]
    \textbf{Ascending low-friction inclines} & $-3.52_{\pm0.4}$  & $-3.58_{\pm0.4}$ & $-3.50_{\pm0.4}$ & $-3.62_{\pm0.5}$ & $-3.65_{\pm0.6}$ & $\boldsymbol{-3.25_{\pm0.3}}$ \\ 

    \bottomrule
  \end{tabular}
  }
\vspace{-0.2cm}
\end{table*}

\vspace{-0.1cm}
Furthermore, hyperparameter selection is performed via systematic grid search over predefined ranges, and the configuration that yields the highest performance on validation rollouts is used for all reported results.
Specifically, we perform a hyperparameter sweep over the learning rates in the range $(10^{-5}, 10^{-3})$ and the forgetting factor in the range $(0.9,0.999)$. Furthermore, we set the batch size to $64$, the MPC horizon to $25$, and the discount factor to $0.99$. In addition, we use an entropy lower bound of $0.2$, a trust region bound of $0.01$, a contraction factor of $0.8$, and $\epsilon = 10^{-4}$.

\noindent
Additionally, all baseline algorithms are tuned via a limited grid search around the default settings provided in their original implementations to ensure fair comparisons. Specifically, for SAC, we use the Adam optimizer with a learning rate of $3\times10^{-4}$, discount factor $\gamma=0.99$, replay buffer size $10^6$, batch size $256$, soft update coefficient $\tau=0.005$, and a neural network comprising two hidden layers of 256 units each. TRPO is configured with a maximum KL divergence $\delta=0.01$, discount factor $\gamma=0.99$, $10$ conjugate gradient steps, and a line search contraction factor of $0.8$. Finally, for Online-DT, we adopt the same model architecture as our method, with context length $K=20$, learning rate $1\times10^{-4}$, weight decay $0.001$, discount factor $\gamma=0.99$, and batch size $256$.


\subsection{Main Results and Discussions}

Table \ref{table:result}, and Figures \ref{fig:robot_neven_terrain}--\ref{fig:robot_inclined_surface} summarize our experimental results on the Unitree AlienGo quadruped across three challenging locomotion tasks. Overall, the MPC-guided DT trained with the proposed Riemannian (curvature-aware) optimizer significantly outperforms classical RL baselines (TRPO, SAC), Online-DT, and our own ablated variants using Euclidean optimizers (Ours-SGD, Ours-Adam). 
Higher average total reward, better sample efficiency, and faster convergence rates evidence this superior performance.

\noindent
As illustrated in the left plots of Figure \ref{fig:robot_neven_terrain} (Locomotion on uneven terrains), Figure \ref{fig:robot_stair_climbing} (Stair climbing), and Figure \ref{fig:robot_inclined_surface} (Ascending low-friction incline), our method achieves higher average total reward with fewer environment steps compared to TRPO, SAC, and Online-DT.
MPC in our framework provides real-time, locally optimal predicted trajectories that guide the decision transformer learning and effectively minimize the need for environment interaction, and reduce the sample complexity.
\noindent
Another key factor contributing to these improvements is the efficient Riemannian optimization strategy. The right plots in Figures \ref{fig:robot_neven_terrain}, \ref{fig:robot_stair_climbing}, and \ref{fig:robot_inclined_surface}, depicting the training loss across iterations, clearly demonstrate the faster convergence of our Riemannian approach compared to traditional Euclidean optimizers like SGD and Adam. This faster convergence is attributed to the Riemannian method's ability to capture the intrinsic geometry of the loss manifold and perform curvature-aware updates, which traditional gradient-based methods often fail to account for.

To complement the summary statistics in Table \ref{table:result}, we performed \emph{Welch's $t$-test} and \emph{Cohen's effect size ($d$)} analysis to assess the significance of performance differences. In the uneven terrain task, our algorithm consistently outperforms TRPO, SAC, and Online-DT, yielding very large $t$-statistics and extremely small $p$-values, indicative of highly significant differences. Specifically, compared to TRPO, our method achieves $t = 20.32$ ($p = 3.60 \times 10^{-8}$) with a Cohen’s $d$ of $12.85$. Against SAC, the results are $t = 7.12$ ($p = 1.00 \times 10^{-4}$, $d = 4.50$), while against Online-DT they are $t = 4.34$ ($p = 3.43 \times 10^{-3}$, $d = 2.75$).


\noindent
In the stair climbing scenario, the statistical analysis indicates a significant difference only in comparison to Online-DT, with $p = 3.63 \times 10^{-2}$. However, across all pairwise comparisons with TRPO, SAC, and Online-DT, the effect sizes are consistently large ($d > 0.8$), suggesting substantial performance gains of our method, even in cases where the $p$-values do not reach conventional thresholds of statistical significance.
%
%
%
\noindent
In the low-friction incline scenario, none of the pairwise comparisons reach conventional levels of statistical significance ($p < 0.05$). Nevertheless, the corresponding Cohen’s $d$ values are all large.
%
%
%
\noindent
Our analysis further shows that the MPC-guided algorithm with Riemannian learning consistently attains large effect sizes relative to the same algorithm trained with SGD or Adam, most notably in the uneven terrain locomotion. Although these improvements are not always statistically significant due to comparatively high $p$-values, they nonetheless highlight the substantial advantage of the Riemannian approach over its Euclidean counterparts.

\section{Conclusions} 

In this work, we proposed a hybrid framework that seamlessly integrates model-based MPC with model-free RL within a sequence modeling paradigm, using the Decision Transformer. To address the challenges arising from the highly non-convex optimization landscape of transformer-based policies, we employed an efficient Riemannian optimization strategy, which enables curvature-aware updates that improve numerical optimization stability and convergence rate. We evaluated the proposed approach on the Unitree AlienGo quadruped robot across three challenging locomotion scenarios: traversing uneven terrains, climbing stairs, and ascending low-friction inclined surfaces.
The results demonstrate that our method consistently outperforms state-of-the-art baselines, achieving higher mean total reward, superior sample efficiency, and faster convergence.

\section*{Acknowledgment}
\label{sec:acknowledgment}

This work was supported in part by the Engineering and Physical Sciences Research Council (EPSRC) through the AI Hub in Generative Models [grant number EP/Y028805/1].


\bibliographystyle{IEEEtran}
\bibliography{References.bib}

\end{document}